\documentclass[conference]{IEEEtran}
\IEEEoverridecommandlockouts

\usepackage[table,xcdraw]{xcolor}
\usepackage{colortbl}
\usepackage{algorithm}

\usepackage{cite}
\usepackage{amsmath,amssymb,amsfonts}
\usepackage{algorithmic}
\usepackage{graphicx}
\usepackage{textcomp}
\usepackage{xcolor}
\def\BibTeX{{\rm B\kern-.05em{\sc i\kern-.025em b}\kern-.08em
    T\kern-.1667em\lower.7ex\hbox{E}\kern-.125emX}}
\begin{document}

\title{RP-CATE: Recurrent Perceptron-based Channel Attention Transformer Encoder for Industrial Hybrid Modeling

}

\author{
\IEEEauthorblockA{\large{Haoran Yang\textsuperscript{1}}\thanks{First author. Email: 22332111@zju.edu.cn}, Yinan Zhang\textsuperscript{1}\IEEEauthorrefmark{2}\thanks{Corresponding author. Email: zhangyinan@zju.edu.cn}, Wenjie Zhang\textsuperscript{2}, Dongxia Wang\textsuperscript{1}, Peiyu Liu\textsuperscript{1}, \\Yuqi Ye\textsuperscript{1}, Kexin Chen\textsuperscript{1}, Wenhai Wang\textsuperscript{1}\IEEEauthorrefmark{2}\thanks{Corresponding author. Email: zdzzlab@zju.edu.cn}}
\\
\IEEEauthorblockA{
\large{\textsuperscript{1}\textit{Department of Control Science and Engineering, Zhejiang University}}\\
\textsuperscript{2}\textit{Risk Department, Binance}
}

}

\maketitle

\begin{abstract}
Nowadays, industrial hybrid modeling which integrates both mechanistic modeling and machine learning-based modeling techniques has attracted increasing interest from scholars due to its high accuracy, low computational cost, and satisfactory interpretability. Nevertheless, the existing industrial hybrid modeling methods still face two main limitations. First, current research has mainly focused on applying a single machine learning method to one specific task, failing to develop a comprehensive machine learning architecture suitable for modeling tasks, which limits their ability to effectively represent complex industrial scenarios. Second, industrial datasets often contain underlying associations (e.g., monotonicity or periodicity) that are not adequately exploited by current research, which can degrade model's predictive performance. To address these limitations, this paper proposes the \textbf{R}ecurrent \textbf{P}erceptron-based \textbf{C}hannel \textbf{A}ttention \textbf{T}ransformer \textbf{E}ncoder (RP-CATE), with three distinctive characteristics: (\romannumeral 1) We developed a novel architecture by replacing the self-attention mechanism with channel attention and incorporating our proposed Recurrent Perceptron (RP) Module into Transformer, achieving enhanced effectiveness for industrial modeling tasks compared to the original Transformer. (\romannumeral 2) We proposed a new data type called Pseudo-Image Data (PID) tailored for channel attention requirements and developed a cyclic sliding window method for generating PID. (\romannumeral 3) We introduced the concept of Pseudo-Sequential Data (PSD) and a method for converting industrial datasets into PSD, which enables the RP Module to capture the underlying associations within industrial dataset more effectively. An experiment aimed at hybrid modeling in chemical engineering was conducted by using RP-CATE and the experimental results demonstrate that RP-CATE achieves the best performance compared to other baseline models.
\end{abstract}


\section{Introduction}
Industrial modeling plays an important role in optimizing the design, operation, and maintenance of industrial systems. By developing mathematical or computational representations of these systems, industrial modeling facilitates the analysis of complex processes, predicts outcomes under various scenarios, and identifies optimal strategies for achieving desired objectives \cite{scheidegger2018introductory}. As the era of Industry 4.0 unfolds, industries face increasing pressure to enhance efficiency, reduce costs, and mitigate environmental harm \cite{borowski2021digitization}, making the development of robust and accurate models more critical than ever.

Most previous work in industrial modeling has focused on the development of mechanistic models. These models leverage domain knowledge and industry-specific mechanisms to rigorously represent industrial processes \cite{velten2024mathematical}. However, mechanistic models typically come with high modeling costs and significant computational efforts \cite{mcbride2020hybrid,elsheikh2023control}. Moreover, their effectiveness is limited in complex industrial systems where underlying mechanisms are either poorly understood or difficult to accurately express. These limitations restrict the broader application of mechanistic models in industrial modeling. On the other hand, machine learning (ML) techniques are advancing at an unprecedented pace \cite{sarker2021machine}. The structures of machine learning models have become increasingly complex and powerful, allowing data-driven models to break free from the constraints of mechanistic models, which rely heavily on complex prior knowledge and mechanistic formulas. As a matter of fact, data-driven techniques have already been employed in industrial modeling for several years \cite{bertolini2021machine,sarker2022ai}. However, these data-driven models are often doubted for its lack of interpretability and scalability \cite{carter2023review}. In this context, industrial hybrid modeling was developed to combine the strengths of both mechanistic and data-driven models, enabling the established models to achieve acceptable interpretability while maintaining low modeling costs \cite{sansana2021recent}. However, existing industrial hybrid modeling methods still suffers from two limitations.

Firstly, existing research has only utilized a single machine learning method for industrial modeling and has not proposed a machine learning architecture applicable to modeling tasks. For instance, Chen and Ierapetritou applied basic machine learning methods like Artificial Neural Networks (ANNs) and Support Vector Regression (SVR) in a hybrid modeling framework to model the Continuous Stirred Tank Reactor (CSTR) \cite{chen2020framework}. Similarly, Li et al. used a surrogate model based on deep Convolutional Neural Networks (CNNs) to model the degradation process of bandsaw blades \cite{li2020novel}. Liu et al. designed a fault diagnosis model based on CNN for electric machine \cite{liu2016dislocated}. Su et al. developed a model for predicting the critical properties of chemicals using machine learning methods such as Deep Neural Networks (DNNs) and Long Short-Term Memory (LSTM) \cite{su2019architecture}. Moreover, the aforementioned ANNs, SVR, CNNs, DNNs and LSTM are relatively outdated compared to the advanced methods, which may further degrade the model's predictive performance.

Second, industrial datasets often contain underlying associations (e.g. monotonicity or periodicity) that are not sufficiently exploited by current research \cite{liu2016dislocated,chen2020framework,li2020novel}. These underlying associations fundamentally stem from the intrinsic mechanisms inherent in the modeling process itself, specifically embodied in the mathematical formulas or equations of its mechanistic model (For a more intuitive explanation of underlying associations, see Definition 1). Undoubtedly, uncovering these underlying associations, such as monotonicity, and incorporating these learned patterns into our prediction process would enhance the predictive performance of the model.

To address these limitations, this paper proposes a novel machine learning-based modeling architecture RP-CATE by improving Transformer architecture. In order to facilitate the model to learn the underlying associations within industrial datasets, we introduced the concept of PSD and designed a PSD Module that transforms input data into PSD. Subsequently, a Recurrent Perceptron (RP) Module was designed to process the PSD, which allowed for learning the underlying associations within industrial datasets. Next, we introduced a Channel Attention Module, which assigns attention weights to different data features based on their importance to the prediction targets. After processing through these modules, the output was passed to a Feed-Forward Module for further nonlinear transformation. The final output of RP-CATE was then generated through the Prediction Module. Similar to the Transformer architecture, the core components of RP-CATE, RP Module, Channel Attention Module, and Feed-Forward Module can be repeated any number of times based on practical requirements. Experimental results demonstrate that RP-CATE achieves the best performance in acentric factor prediction, outperforming all other benchmarks and showing an 84.15\% improvement in the MIR (Model Improvement Rate) metric compared to the mechanistic model. To summarize, our main contributions are as follows.
\begin{itemize}
    \item We proposed a novel machine learning-based modeling architecture RP-CATE by improving Transformer architecture. 
    \item We developed the PSD Module and RP Module, which effectively learns and leverages the underlying associations within industrial datasets, thereby enhancing the model’s predictive performance.
    \item Considering that different data features have varying impacts on the target value, we enhanced the channel attention mechanism and integrated the improved version into the proposed RP-CATE architecture to assign appropriate attention weights to the features.
    \item We conducted an experiment in chemical engineering to validate the effectiveness of the proposed architecture and demonstrated its superiority over other baselines. The results show that our method significantly outperforms all baseline methods.
\end{itemize}

\section{Methodology}
\subsection{Problem Formulation}

In modeling industrial processes, one of the most prevalent and challenging problems is the bias between the output of mechanistic models and the actual output observed in real-world scenarios \cite{mcbride2020hybrid,elsheikh2023control}. Therefore, we consider using data-driven models to correct these bias. By integrating the data-driven model with the mechanistic model in a parallel configuration, we aim to develop a hybrid industrial model that more accurately reflects the true behavior of the object being modeled. Thus, the dataset used in this study consists of \(X\) and \(Y\), where \(X\) represents the actual input of the modeling object, and \(Y\) represents the bias between the actual output of the modeling object and the output of its mechanistic model. \(X=\{x_{1},x_{2},...,x_{m}\}, x_{i} \in \mathbb{R}^{n}; Y=Y_{true}-Y_{me}=\{y_{1},y_{2},...,y_{m}\}, y_{i} \in \mathbb{R}\). Here, \(Y_{true}\) denotes the actual output of the modeling object, and \(Y_{{me}}\) denotes the output of the purely mechanistic model. Furthermore, we assume that $X$ contains \(m\) samples, each with \(n\) features, and that $X$ is non-strictly-sequential data. 

Whether in a mechanistic model or a data-driven model, the input is consistently denoted as \(X\). The various features of $X$ correspond to different independent variables in the mechanistic model, where each independent variable is associated with a coefficient of varying magnitude. From the perspective of data-driven modeling, these varying coefficients represent the differential impact that each feature has on the target values. Therefore, different features in $X$ have differential impact on the target values.

\subsubsection{Definition 1 Non-Strictly-Sequential Data (NSSD)} \textit{NSSD refers to data that does not inherently follow a strictly sequential form, but exhibit underlying associations such as monotonicity or periodicity.}

\textit{Example: Industrial data often contains some underlying associations, which fundamentally arise from the intrinsic mechanisms inherent to the process being modeled. Let’s denote this mechanism by \(f\). For instance, industrial data might satisfy the condition \(f(x) = f(x+T)\), indicating a periodic pattern where \(T\) represents the period. Alternatively, the data might exhibit monotonicity, represented by \(\frac{\partial f}{\partial x} > 0\) for increasing trends or \(\frac{\partial f}{\partial x} < 0\) for decreasing trends.
}

\begin{figure*}[t]
\centering
\includegraphics[width=0.95\textwidth]{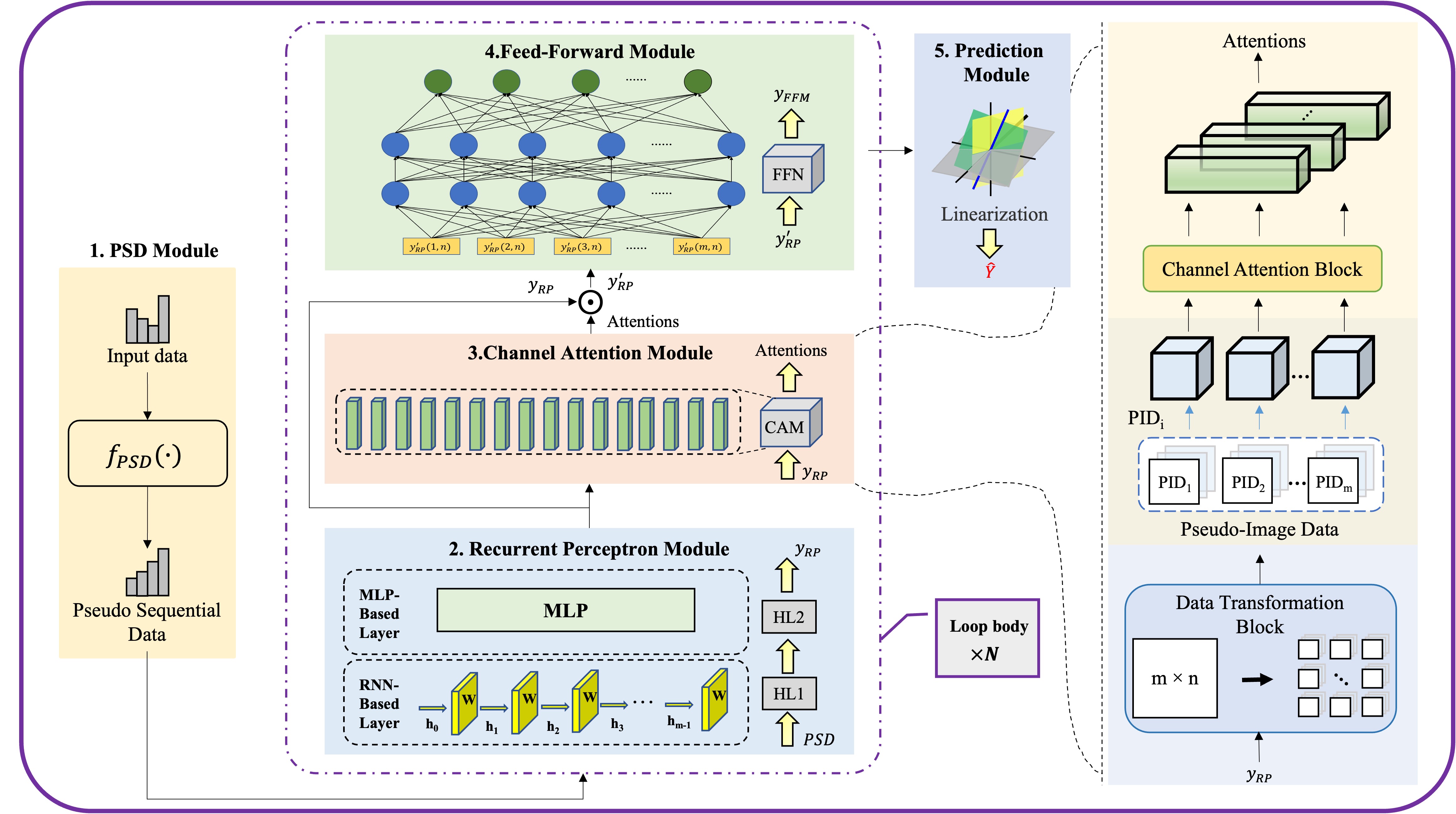} 
\caption{Overview of the RP-CATE framework.}
\label{fig-1}
\end{figure*}

\subsubsection{Definition 2 Pseudo-Sequential Data (PSD)}
\textit{PSD is a data structure derived from NSSD. It is obtained by sorting the sample data according to the numerical values of a given feature \( x' \), transforming NSSD into a form where the data exhibits an ordered arrangement. Mathematically, $PSD$ is defined as \({PSD}={sort}(X; x^{\prime}) \), where \( {PSD}_i \in {PSD}, i = 1, 2, \dots, m \).}

\textit{Example: The input $X$, which is NSSD, each record $x_i$ represents a sample, whereas in $PSD$, each record \({PSD_i}\) indicates the input to the subsequent module at time step $t=i$.}

\subsection{Overview}
Fig.~\ref{fig-1} depicts our proposed RP-CATE architecture, which consists of five main modules. (1) \textbf{PSD Module.} In order to make the model learn the underlying associations between different samples in the input dataset more efficiently, we developed PSD Module to convert the original industrial data (NSSD) into PSD. (2) \textbf{Recurrent Perceptron Module}. We proposed the Recurrent Perceptron (RP) to capture the associations between the current input \(\text{PSD}_{i}\) and the previous input \(\text{PSD}_{i-1}\) from the PSD. (3) \textbf{Channel Attention Module}. Given that different features in $X$ have differential impact on prediction outcomes, this module was developed to assign appropriate weights to different features according to their importance. (4) \textbf{Feed-Forward Module}. Moreover, we introduced a feed-forward module to enhance the model’s ability to fit nonlinear relationships. (5) \textbf{Prediction Module}. Finally, the data were fed into this module to generate the predicted label values. The modules from (2) to (4) can be iterated as needed to enhance model performance.

\subsection{PSD Module}
The input \(X \in \mathbb{R}^{m \times n}\) is NSSD, where underlying associations between different samples may exist as outlined in Definition 1. While these associations can theoretically be captured by an Recurrent Neural Network (RNN) layer \cite{elman1990finding,xiao2017modeling}, the non-strictly-sequential characteristic in NSSD makes it challenging for the model to learn these associations effectively. Directly applying RNN to NSSD may decrease the model's efficiency and potentially affect its predictive performance. To address this, we proposed the concept of PSD, with the transformation process as follows:

\begin{equation}
    PSD={f}_{PSD}(X;{x}^{\prime})=sort(X;{x}^{\prime})\in \mathbb{R}^{m\times n}.
    \label{equa-01}
\end{equation}

\noindent The resulting \(PSD\) contains $m$ records \(\{PSD_1,PSD_2,...,PSD_m\}\), and each record \({PSD}_i \in \mathbb{R}^n\) is no longer treated as an independent sample but instead becomes the input to the subsequent Recurrent Perceptron Module at different time steps.

\subsection{Recurrent Perceptron Module}

The proposed RP Module is used to process PSD by capturing associations between the current input $\text{PSD}_i$ and the previous input $\text{PSD}_{i-1}$ to learn the underlying associations between industrial data, thereby improving the model’s predictive performance. In this module, the Recurrent Perceptron combines the architectures of RNN and Multilayer Perceptron (MLP). The designed RP has two layers: the first layer (HL1) is a hidden layer based on the RNN architecture, designed to learn the associations between inputs at different time steps. The second layer (HL2) is an output layer based on the MLP architecture, which performs further nonlinear transformations on the output from HL1. The forward propagation process is outlined as follows:

\begin{equation}
\begin{aligned}
    {h}_{i}&=RNN(PSD_{i},h_{i-1};U,W,b_{HL1})
         \\&=\sigma(PSD_{i}U+h_{i-1}W+b_{HL1}),
\end{aligned}
\label{equa-02}
\end{equation}

\begin{equation}
    \begin{aligned}             {y}_{RP,i}&=MLP(h_{i};V,W_{HL2},b,b_{HL2})\\&=\sigma(\sigma(h_{i}V+b_{HL2})W_{HL2}+b),
    \label{equa-03}
    \end{aligned}
\end{equation}

\noindent where \(U, W, V, W_{HL2}, b_{HL1}, b_{HL2}\), and $b$ are the parameters to be trained, \(\sigma\) represents the sigmoid activation function, $i$ represents the current time step, ranging from 1 to $m$, \(h_0\) is initialized to 0, and \(y_{RP,i}\) denotes the outputs of RP Module at time step $i$. Thus, this module can be defined as follows:

\begin{equation}
\scalebox{0.85}{$
        {y}_{RP}=f_{RP}(PSD;U,V,W,W_{HL2},b_{HL1},b_{HL2},b)\in
        \mathbb{R}^{m\times n}.
    \label{equa-04}
$}
\end{equation}

\begin{figure}[t]
\centering
\includegraphics[width=0.9\columnwidth]{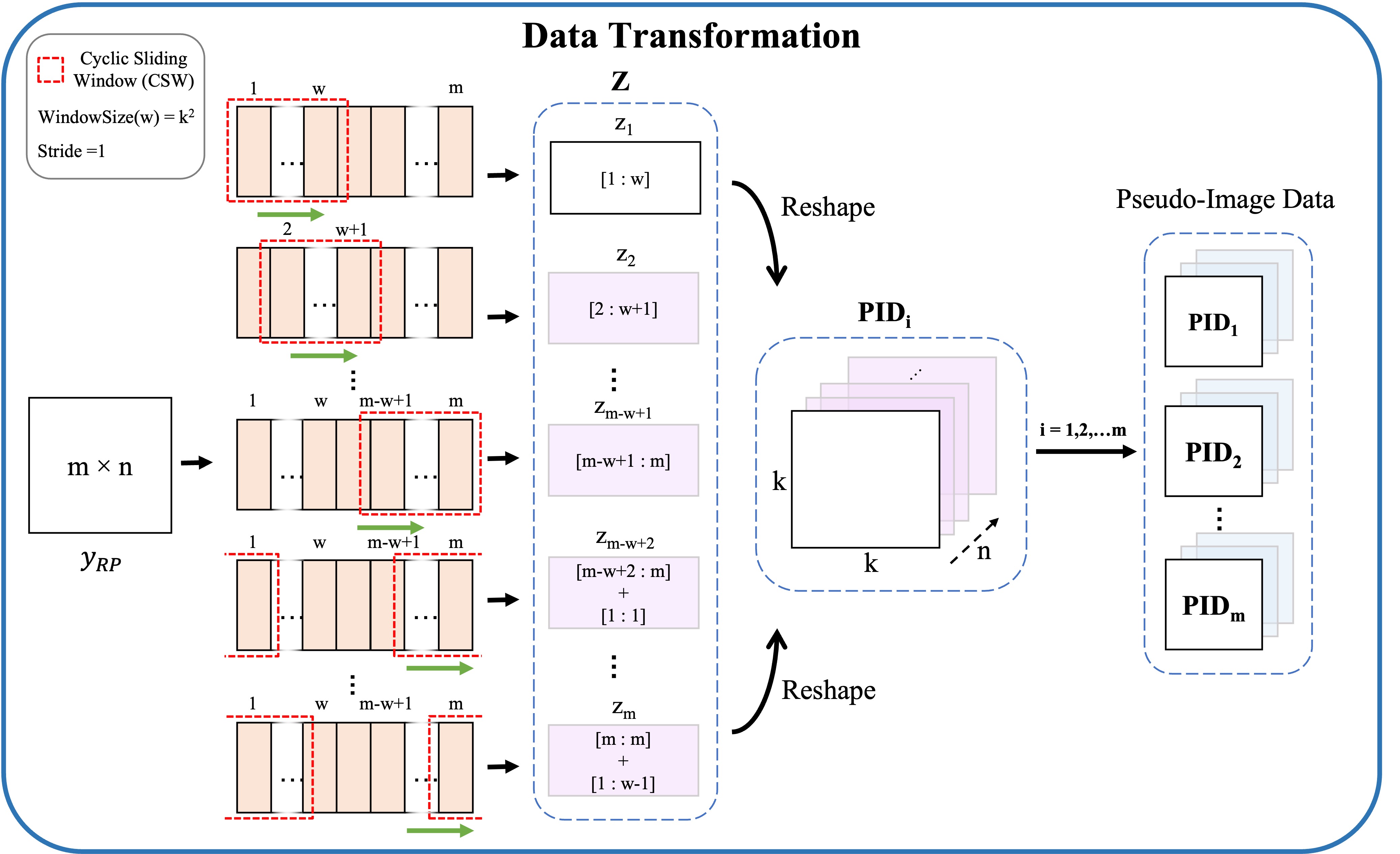} 
\caption{The process of data transformation. The red dashed box in the figure represents the Cyclic Sliding Window (CSW), with a window size \(w\) set to \(k^2\) and a sliding step size of 1. For each sliding step, the CSW produces a data set \(z_i\). After reaching the \(m - w + 1\)-th step, there will be gaps within the window. These gaps will cyclically wrap around to the beginning of the input, filling the gaps in turn until the window has traversed all the data from left to right. Ultimately, \(CSW(.)\) will generate the data set \(z_i\) with the same number of elements as the original data, where \(i=\{1,2,...,m\}\).}
\label{fig-2}
\end{figure}

\subsection{Channel Attention Module}

According to the previous discussion, if the model can adequately account for the varying importance of different features and incorporate this into its prediction process, it will significantly enhance the model’s performance. To achieve this purpose, we developed the Channel Attention Module to assign appropriate attention weights to different features. 

As illustrated in Fig.~\ref{fig-1}, the Channel Attention Module comprises two blocks. The first is the Data Transformation Block. Since the data structure output by RP Module cannot be directly processed by the subsequent Channel Attention Block, it needs to be processed firstly. To address this, we introduce the concept of Pseudo-Image Data (PID), a novel data representation, specifically designed for the Channel Attention Block. This block aims to transform the module’s input into the PID. The second block is the Channel Attention Block, which can process PID and generate the corresponding attention weights for this module’s input.

\subsubsection{Definition 3 Pseudo-Image Data (PID)} \textit{PID is a data representation similar to the RGB data structure of image. Given the input data \(X \in \mathbb{R}^{m \times n}\), after applying Cyclic Sliding Window (\(CSW(.)\)) with window size w produces \(Z=\{z_{1},z_{2},...,z_{m}\}, z_{i}\in \mathbb{R}^{w \times n}\), i.e., \(Z = {CSW}(X) \in \mathbb{R}^{m \times w \times n}\). Reshaping \(Z\) results in \(PID\), i.e., \({PID} = {reshape}(Z) \in \mathbb{R}^{m \times \sqrt{w} \times \sqrt{w} \times n}\). Note that the window size w  must be a perfect square.}

\subsubsection{Data Transformation Block}
As shown in the Fig.~\ref{fig-2}, we first set the window size \(w\) for the $CSW$ based on the requirements of the specific task. Next, the input data is processed using this \(CSW(.)\), resulting in the corresponding $PID$. The detailed process is as follows:

\begin{equation}
    \begin{aligned}
        PID &= f_{PID}(y_{RP}; w) \\
        &= reshape({CSW}(y_{RP})) \in \mathbb{R}^{m \times k \times k \times n}, \\
        &  s.t. \quad  w = k^{2},\ k \in \mathbb{Z}^{+}
    \end{aligned}
\end{equation}

\subsubsection{Channel Attention Block}
The \(PID\) obtained through the above process can be represented as \({PID} = \{{PID}_1, {PID}_2, \ldots, {PID}_m \} \in \mathbb{R}^{m \times k \times k \times n}\), where each sample is \({PID}_i \in \mathbb{R}^{k \times k \times n}\). The execution process of channel attention is illustrated in the Fig.~\ref{fig-3}. First, all the sample \(PID_{i}\) undergoes global max pooling \(GMP(.)\) and global average pooling \(GAP(.)\). The results are then squeezed to remove redundant dimensions. Next, the resulting vectors are passed through separate feed-forward network \(FFN(.)\) and summed. The summed result is then processed through a sigmoid activation function for nonlinear transformation. Finally, the output is processed through a softmax layer to obtain the attention weights for \(PID_{i}\), as shown in the following formulas:

\begin{figure}[t]
\centering
\includegraphics[width=0.9\columnwidth]{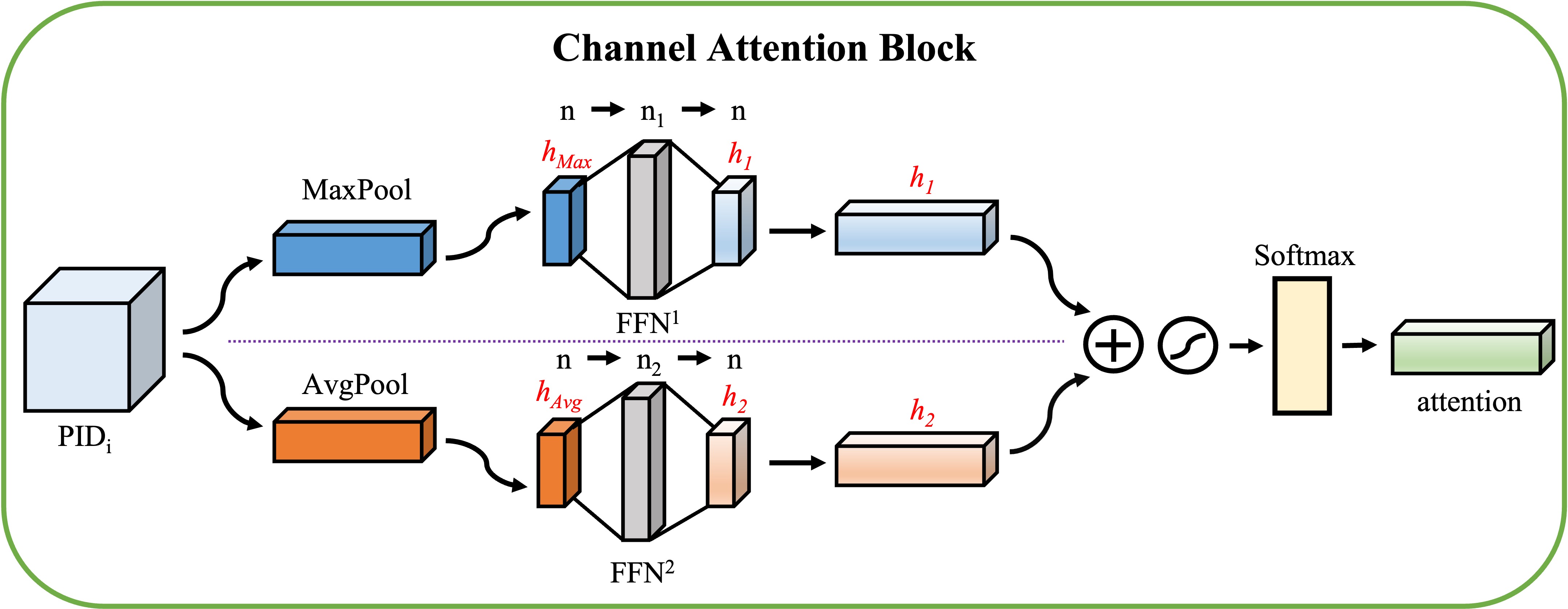} 
\caption{Diagram of Channel Attention Block. The figure demonstrates the process of generating attention using a sample ${PID}_i$ from $PID$. In this process, the red italicized text represents intermediate variables. Specifically, \(h_{Max} \in H_{Max}, h_{Avg} \in H_{Avg}, h_1 \in H_1, h_2 \in H_2\) and \(attention \in Attentions\)}
\label{fig-3}
\end{figure}

\begin{equation}
    H_{Max} = squeeze(GMP(PID))\in \mathbb{R}^{m\times n},
    \label{equa-06}
\end{equation}

\begin{equation}
    H_{Avg} = squeeze(GAP(PID))\in \mathbb{R}^{m\times n},
    \label{equa-07}
\end{equation}

\begin{equation}
    H_1 = FFN^1(H_{Max})\in \mathbb{R}^{m \times n},n\xrightarrow{} n^1 \xrightarrow{}n,
    \label{equa-08}
\end{equation}

\begin{equation}
    H_2 = FFN^2(H_{Avg})\in \mathbb{R}^{m \times n},n\xrightarrow{} n^2 \xrightarrow{}n,
    \label{equa-09}
\end{equation}

\begin{equation}
\begin{aligned}
        Attentions &= softmax(\sigma(H_1+H_2))\in \mathbb{R}^{m \times n},
        \\& s.t. \quad n^1=n^2>n,
\end{aligned}
\label{equa-10}
\end{equation}

\noindent where \({FFN}^1\) and \({FFN}^2\) are both two-layer MLP with ReLU activation, with configurations of \(\{n_1, n\}\) and \(\{n_2, n\}\), respectively. The entire channel attention module can be defined as follows:

\begin{equation}
    Attentions = f_{CAM}(y_{RP};w)\in \mathbb{R}^{m \times n}.
    \label{equa-11}
\end{equation}

\subsection{Feed-Forward Module}
The Feed-Forward Module is designed to perform nonlinear transformations. It primarily consists of a feed-forward network, specifically an MLP with three layers and sigmoid activation function. The input to this module is \({y}_{RP}^{\prime}\), the attention-weighted \(y_{RP}\):

\begin{equation}
    y_{RP}^{\prime}=y_{RP}\odot Attentions \in \mathbb{R}^{m \times n},
    \label{equa-12}
\end{equation}

\noindent where \(\odot\) denotes the Hadamard product. The resulting \(y_{RP}^{\prime}\) is then input into the \(FFN^3\) with a configuration of \(\{n^3,n^4,n\}\), producing the output of the Feed-Forward Module, as defined by the following formula:

\begin{equation}
\begin{aligned}
        y_{FFM} = f_{FFM}&(y_{RP}^{\prime})=FFN^3(y_{RP}^{\prime}) \in \mathbb{R}^{m \times n},
        \\&n\xrightarrow{}n^3\xrightarrow{}n^4\xrightarrow{}n,
        \\& s.t. \quad n^3>n,n^4>n.
\end{aligned}
\label{equa-13}
\end{equation}

\subsection{Prediction Module and Training}
The Prediction Module is a layer without nonlinear activation functions. It applies linear transformation to \(y_{FFM}\), obtained after processing through the four modules described above, to produce the model’s final predicted values.

\begin{equation}
    \widehat{Y}=f_L(y_{FFM})=y_{FFM}W_{L}+b_L\in \mathbb{R}^{m \times 1},
\end{equation}

\noindent where \(W_{L} \in \mathbb{R}^{n \times 1}\) and \(b_{L} \in \mathbb{R}^{m \times 1}\) are the trainable parameters of the Prediction Module. Additionally, \(y_{FFM}\) can be used to establish a residual connection with the initially obtained $PSD$, and then be fed back into the RP Module for further forward propagation. This process can be repeated any number of times as needed for the specific task. The completed forward propagation process of RP-CATE is described in Algorithm \ref{algorithm}.

\begin{algorithm}[tb]
\caption{The forward propagation of RP-CATE}
\label{algorithm}
\raggedright
\textbf{Input}: \( X = \{x_1, x_2, \ldots, x_m\} \in \mathbb{R}^{m \times n} \)\\
\textbf{Parameter}: \( {x^{\prime}, N, w} \) \\
\textbf{Output}: \( \widehat{Y} = \{\widehat{y_1}, \widehat{y_2}, \ldots, \widehat{y_m}\} \in \mathbb{R}^{m \times 1} \)
\vspace{-1em}
\begin{algorithmic}[1] 
\STATE Let $i=0$.
\STATE \(PSD=f_{PSD}(X;x^{\prime})=sort(X;x^{\prime}),I=PSD\)
\WHILE{\(i<N\)}
\IF {\(i>0\)}
\STATE $PSD=PSD+I$
\ENDIF
\STATE \(y_{RP}=f_{RP}(PSD)\)
\STATE \(Atts=f_{CAM}(y_{RP};w),y_{RP}^{\prime}=Atts\odot y_{RP}\)
\STATE \(y_{FFM}=f_{FFM}(y_{RP}^{\prime})\)
\STATE \(\widehat{Y}=f_L(y_{FFM})\)
\STATE $i=i+1$
\ENDWHILE
\STATE \textbf{return} $\widehat{Y}$
\end{algorithmic}
\end{algorithm}

After completing the forward propagation process, we obtain the predicted labels \(\widehat{Y}\) for the training set. The trainable parameters within each module of RP-CATE are optimized by minimizing the following loss function:

\begin{equation}
    \mathcal{L}=\frac{1}{m}\displaystyle\sum_{i=1}^{m}(y_i-\widehat{y_i})^2+\lambda\parallel \varTheta  \parallel_2,
    \label{equa-15}
\end{equation}

\noindent where \(y_i \in Y\) and \(\hat{y}_i \in \hat{Y}\) denote the actual label and the predicted label for the i-th training sample, respectively. The term  \(\parallel \varTheta  \parallel_2\) serves as a regularization term to prevent model overfitting, while \(\lambda\) controls the regularization strength.

\section{Experiments}

In this section, we conduct experiments to address the following four questions: 

\begin{itemize}
    \item \textbf{RQ1:} How does varying the number of repetitions N in the model’s loop body affect the prediction results? Is it superior to a Transformer with the same configuration?
    \item \textbf{RQ2:} How is the overall performance of RP-CATE compared with established methods?
    \item \textbf{RQ3:} Does each module of RP-CATE contribute to improving the predictive performance?
    \item \textbf{RQ4:} What insights on the different features correlation can we get through visualizing the attention map.
\end{itemize}

\subsubsection{Modeling Object}
Petroleum is a complex mixture that exhibits non-ideal behavior under the high-temperature and high-pressure conditions typical of refining processes. These deviations can lead to significant inaccuracies in simulation results that rely on idealized models. To account for such discrepancies, the acentric factor—a dimensionless parameter that quantifies the deviation of real gases from ideal gas behavior—is introduced to correct the discrepancies between idealized and non-idealized behaviors \cite{adewumi2013acentric}. Among the various methods for calculating acentric factors, the Lee-Kesler (LK) method is widely used due to its simple structure and broad applicability \cite{lee1975generalized,boublia2023critical}. However, the accuracy of the LK method’s results significantly deteriorates when the input variables fall outside a certain range \cite{abakporo2021general}. To address this limitation, we focus on modeling the acentric factor using a hybrid approach: combining the LK mechanistic model with our proposed RP-CATE data-driven model, which learns to correct the systematic bias in LK’s predictions. The final acentric factor is obtained by adding this correction term to the LK output, thereby improving simulation reliability and overall prediction accuracy.

\subsubsection{Baselines}
To comprehensively evaluate the model performance and compare the differences across various models, we compare the performance of RP-CATE with the mechanistic method and several other baseline approaches.
\textbullet \hspace{1pt} {Lee-Kesler Method} \cite{lee1975generalized}: A mechanistic model calculation method of acentric factor. \textbullet \hspace{1pt} {RBFNN} \cite{banchero2018comparison}: A feed-forward neural network with a hidden layer composed of radial basis neurons and using radial basis functions as the activation functions. \textbullet \hspace{1pt} {LSTM} \cite{zhao2017time,sherstinsky2020fundamentals}, {GRU} \cite{mahjoub2022predicting}: Popular variants of RNNs for sequential prediction. \textbullet \hspace{1pt} {Transformer} \cite{vaswani2017attention}: A novel network architecture based solely on an attention mechanism, eliminating the need for recurrence and convolutions. \textbullet \hspace{1pt} {TCN} \cite{bai2018empirical}: A network capable of taking a sequence of any length and mapping it to an output sequence of the same length. \textbullet \hspace{1pt} {TGCN-S} \cite{kong2023collaborative}: A graph neural network using a graph structure learning module to learn potential intervariable relationships from data. \textbullet \hspace{1pt} {DGDL} \cite{zhu2022dynamic}: A graph neural network using a dynamic graph to realize adaptive learning and automatic inference. \textbullet \hspace{1pt} {RADA} \cite{chen2024residual}: A residual-aware deep attention graph convolutional network which uses a residual-aware connection module to reduce data uncertainty and alleviate over-smoothing.

\renewcommand{\arraystretch}{1.5}

\begin{table*}[t]
\centering
\large
\caption{The comparative experiment between RP-CATE and Transformer. The best results between RP-CATE and Transformer are marked with \textsuperscript{*}. The best results for RP-CATE are in bold, and the best results for Transformer are underlined.}
\resizebox{\textwidth}{!}{%
\begin{tabular}{c|c|cc|cc|cc|cc|cc|
>{\columncolor[HTML]{EFEFEF}}c 
>{\columncolor[HTML]{EFEFEF}}c }
\hline

  \multicolumn{1}{l|}{} &
   &
  \multicolumn{2}{p{4cm}|}{\centering MAE} &
  \multicolumn{2}{p{4cm}|}{\centering RMSE} &
  \multicolumn{2}{p{4cm}|}{\centering AER(\%)} &
  \multicolumn{2}{p{4cm}|}{\centering \#Err\textless{}1\%} &
  \multicolumn{2}{p{4cm}|}{\centering \#Err\textgreater{}5\%} &
  \multicolumn{2}{c}{\cellcolor[HTML]{EFEFEF} \centering MIR(\%)}   \\ \cline{3-14} 
  
\multicolumn{1}{c|}{\centering Window Size(w)} &
  Repetitions(N) &
  \multicolumn{1}{p{2cm}|}{\centering RP-CATE} &
  Transformer &
  \multicolumn{1}{p{2cm}|}{\centering RP-CATE} &
  Transformer &
  \multicolumn{1}{p{2cm}|}{\centering RP-CATE} &
  Transformer &
  \multicolumn{1}{p{2cm}|}{\centering RP-CATE} &
  Transformer &
  \multicolumn{1}{p{2cm}|}{\centering RP-CATE} &
  Transformer &
  \multicolumn{1}{p{2cm}|}{\cellcolor[HTML]{EFEFEF} \centering RP-CATE} &
  Transformer  \\ \hline

\textbf{} &
  N=1 &
  \multicolumn{1}{c|}{{0.0104}\textsuperscript{*}} &
  0.0371 &
  \multicolumn{1}{c|}{{0.0126}\textsuperscript{*}} &
  0.0443 &
  \multicolumn{1}{c|}{{1.65}\textsuperscript{*}} &
  4.76 &
  \multicolumn{1}{c|}{{30}\textsuperscript{*}} &
  2 &
  \multicolumn{1}{c|}{{5}\textsuperscript{*}} &
  20 &
  \multicolumn{1}{c|}{\cellcolor[HTML]{EFEFEF}{58.38}\textsuperscript{*}} &
  8.84 \\
 &
  N=2 &
  \multicolumn{1}{c|}{{0.0092}\textsuperscript{*}} &
  0.0372 &
  \multicolumn{1}{c|}{{0.0148}\textsuperscript{*}} &
  0.0543 &
  \multicolumn{1}{c|}{{0.94}\textsuperscript{*}} &
  3.43 &
  \multicolumn{1}{c|}{{50}\textsuperscript{*}} &
  10 &
  \multicolumn{1}{c|}{{0}\textsuperscript{*}} &
  15 &
  \multicolumn{1}{c|}{\cellcolor[HTML]{EFEFEF}{82.48}\textsuperscript{*}} &
  18.04 \\
w=9 &
  N=3 &
  \multicolumn{1}{c|}{{0.0097}\textsuperscript{*}} &
  0.0928 &
  \multicolumn{1}{c|}{{0.0142}\textsuperscript{*}} &
  0.1297 &
  \multicolumn{1}{c|}{{1.54}\textsuperscript{*}} &
  10.82 &
  \multicolumn{1}{c|}{{35}\textsuperscript{*}} &
  2 &
  \multicolumn{1}{c|}{{1}\textsuperscript{*}} &
  47 &
  \multicolumn{1}{c|}{\cellcolor[HTML]{EFEFEF}{64.70}\textsuperscript{*}} &
  -26.83 \\
 &
  N=4 &
  \multicolumn{1}{c|}{{0.0077}\textsuperscript{*}} &
  0.0887 &
  \multicolumn{1}{c|}{{0.0125}\textsuperscript{*}} &
  0.1392 &
  \multicolumn{1}{c|}{{0.73}\textsuperscript{*}} &
  9.20 &
  \multicolumn{1}{c|}{{44}\textsuperscript{*}} &
  3 &
  \multicolumn{1}{c|}{{1}\textsuperscript{*}} &
  39 &
  \multicolumn{1}{c|}{\cellcolor[HTML]{EFEFEF}{76.46}\textsuperscript{*}} &
  -23.03 \\
 &
  N=5 &
  \multicolumn{1}{c|}{{0.0079}\textsuperscript{*}} &
  0.0484 &
  \multicolumn{1}{c|}{{0.0125}\textsuperscript{*}} &
  0.0830 &
  \multicolumn{1}{c|}{{0.84}\textsuperscript{*}} &
  6.32 &
  \multicolumn{1}{c|}{{40}\textsuperscript{*}} &
  16 &
  \multicolumn{1}{c|}{{1}\textsuperscript{*}} &
  22 &
  \multicolumn{1}{c|}{\cellcolor[HTML]{EFEFEF}{71.66}\textsuperscript{*}} &
  17.87 \\ \hline
\textbf{} &
  N=1 &
  \multicolumn{1}{c|}{{\textbf{0.0065}\textsuperscript{*}}} &
  {\underline {0.0279}} &
  \multicolumn{1}{c|}{{{\textbf{0.0110}\textsuperscript{*}}}} &
  {\underline {0.0428}} &
  \multicolumn{1}{c|}{{{\textbf{0.58}\textsuperscript{*}}}} &
  {\underline {2.62}} &
  \multicolumn{1}{c|}{{54}\textsuperscript{*}} &
  9 &
  \multicolumn{1}{c|}{{\textbf{0}\textsuperscript{*}}} &
  {\underline {9}} &
  \multicolumn{1}{c|}{\cellcolor[HTML]{EFEFEF}{88.84}\textsuperscript{*}} &
  {\underline {22.85}} \\
 &
  N=2 &
  \multicolumn{1}{c|}{{0.0067}\textsuperscript{*}} &
  0.0846 &
  \multicolumn{1}{c|}{{0.0113}\textsuperscript{*}} &
  0.1664 &
  \multicolumn{1}{c|}{{0.59}\textsuperscript{*}} &
  7.07 &
  \multicolumn{1}{c|}{{{54}\textsuperscript{*}}} &
  {\underline {12}} &
  \multicolumn{1}{c|}{{ {\textbf{0}\textsuperscript{*}}}} &
  26 &
  \multicolumn{1}{c|}{\cellcolor[HTML]{EFEFEF}{{88.74}\textsuperscript{*}}} &
  -9.92 \\
w=25 &
  N=3 &
  \multicolumn{1}{c|}{{0.0072}\textsuperscript{*}} &
  0.1011 &
  \multicolumn{1}{c|}{{0.0113}\textsuperscript{*}} &
  0.1371 &
  \multicolumn{1}{c|}{{0.80}\textsuperscript{*}} &
  12.32 &
  \multicolumn{1}{c|}{{49}\textsuperscript{*}} &
  1 &
  \multicolumn{1}{c|}{{\textbf{0}\textsuperscript{*}}} &
  46 &
  \multicolumn{1}{c|}{\cellcolor[HTML]{EFEFEF}{82.59}\textsuperscript{*}} &
  -33.27 \\
 &
  N=4 &
  \multicolumn{1}{c|}{{0.0066}\textsuperscript{*}} &
  0.1330 &
  \multicolumn{1}{c|}{{0.0115}\textsuperscript{*}} &
  0.1647 &
  \multicolumn{1}{c|}{{0.59}\textsuperscript{*}} &
  17.11 &
  \multicolumn{1}{c|}{{\textbf{55}\textsuperscript{*}}} &
  1 &
  \multicolumn{1}{c|}{{\textbf{0}\textsuperscript{*}}} &
  51 &
  \multicolumn{1}{c|}{\cellcolor[HTML]{EFEFEF}{\textbf{89.93}\textsuperscript{*}}} &
  -53.71 \\
 &
  N=5 &
  \multicolumn{1}{c|}{{0.0067}\textsuperscript{*}} &
  0.1352 &
  \multicolumn{1}{c|}{{0.0116}\textsuperscript{*}} &
  0.1905 &
  \multicolumn{1}{c|}{{0.61}\textsuperscript{*}} &
  12.90 &
  \multicolumn{1}{c|}{{54}\textsuperscript{*}} &
  1 &
  \multicolumn{1}{c|}{{\textbf{0}\textsuperscript{*}}} &
  53 &
  \multicolumn{1}{c|}{\cellcolor[HTML]{EFEFEF}{88.75}\textsuperscript{*}} &
  -55.12 \\ \hline
\end{tabular}%
}

\label{Table-1}
\end{table*}

\subsubsection{Metrics}
We select the following metrics to give a thorough evaluation of the model performance. (1) MAE (Mean Absolute Error). (2) RMSE (Root Mean Squared Error) \cite{banchero2018comparison}. (3) ARE (Average Relative Error) \cite{su2019architecture}. (4) MIR (Model Improvement Rate): the rate of improvement of the model. (5) \#Err\(<\)1\%: the number of samples with a relative error less than 1\% \cite{su2019architecture}. (6) \#Err\(>\)5\%: the number of samples with a relative error more than 5\% \cite{su2019architecture}. Among these metrics, MIR is a performance indicator we proposed, representing the improvement of the current model compared to the mechanistic model (the mechanistic model has a MIR of 0\%), and it is calculated as follows:

\begin{equation}
\scalebox{0.9}{$
\begin{aligned}
    \text{MIR} &= \left(\frac{1}{2} \times \left(1 - \frac{\sum_{i=1}^{m} \left| \widehat{y_i} + y_{me,i} - y_{true,i} \right|}{\sum_{i=1}^{m} \left| y_{me,i} - y_{true,i} \right|}\right) \right. \\
    &+ \left.\frac{1}{2} \times \left(\frac{\# \text{Err}<1\%  - \#_{me} \text{Err}<1\%}{m - \#_{me} \text{Err}<1\%}\right) \right)\times 100\%
    \label{euqa-16} 
\end{aligned}
$}
\end{equation}

\noindent where, $m$ represents the number of samples in the dataset. \(\widehat{y_i} \in \widehat{Y}\), \(y_{me,i} \in Y_{me}\), and \(y_{true,i} \in Y_{true}\) correspond to the data-driven model predictions, the mechanistic model results, and the actual outputs of the modeling object, respectively. \(\#\text{Err}<1\%\) and \(\#_{me}\text{Err}<1\%\) represent the number of samples with a relative error less than 1\% in the predictions of the hybrid model and the mechanistic model, respectively.

\begin{table*}[t]
\centering
\large
\caption{Overall performance comparison. The best results are in bold and the second-best results are underlined}
\resizebox{\textwidth}{!}{%
\begin{tabular}{c|c|c|c|c|c|
>{\columncolor[HTML]{EFEFEF}}c }
\hline

  \multicolumn{1}{c|}{\centering Model} &
  \multicolumn{1}{p{4cm}|}{\centering MAE} &
  \multicolumn{1}{p{4cm}|}{\centering RMSE} &
  \multicolumn{1}{p{4cm}|}{\centering AER(\%)} &
  \multicolumn{1}{p{4cm}|}{\centering \#Err\textless{}1\%} &
  \multicolumn{1}{p{4cm}|}{\centering \#Err\textgreater{}5\%} &
  \multicolumn{1}{p{4cm}}{\cellcolor[HTML]{EFEFEF} \centering MIR(\%)}

\\ \hline
Lee-Kesler  & 0.0781        & 0.1534       & 5.35 & 17                  & 17                     & 0           \\
RBFNN       & $0.0601\pm0.0258$ & $0.0714\pm0.0288$ & $12.14\pm5.96$    & $4\pm4$                 & $35\pm10$                  & $-3.59\pm13.96$ \\
GRU         & $0.0188\pm0.0870$ & $0.0326\pm0.0150$  & $1.34\pm0.51$     & $34\pm7$                & $3\pm3$                    & $57.76\pm23.70$ \\
LSTM        & $0.0104\pm0.0033$ & $0.0157\pm0.0872$ & $0.94\pm0.11$     & $32\pm8$                & \underline{$0+1$}              & $60.77\pm9.30$  \\
Transformer & $0.0345\pm0.0075$ & $0.0566\pm0.0186$ & $3.94\pm1.85$     & $15\pm6$                & $17\pm8$                   & $25.58\pm7.21$  \\
TCN         & \underline{$0.0086\pm0.0022$} & \underline{$0.0153\pm0.0049$} & \underline{$0.86\pm0.12$}     & \underline{$46\pm8$}                & {\textbf{0}}             & \underline{$78.24\pm8.40$}  \\
TGCN-S    &$0.0192\pm0.0056$    &$0.0250\pm0.0054$    &  $2.78\pm1.23$ &   $22\pm10$&  $8\pm10$ & $55.28\pm26.59$  \\
DGDL        &$0.0261\pm0.0113$   &$0.0323\pm0.0112$    & $2.15\pm2.97$  & $13\pm7$  & $4\pm8$  & $44.96\pm29.19$  \\
RADA        &$0.0143\pm0.0007$    &$0.0194\pm0.0008$    &   $1.86\pm0.45$& $25\pm4$  & $1\pm2$  & $56.39\pm8.69$  \\

\textbf{RP-CATE (Ours)} & {$\textbf{0.0084}\pm\textbf{0.0017}$} & {$\textbf{0.0132}\pm\textbf{0.0020}$} & {$\textbf{0.81}\pm\textbf{0.09}$} & {$\textbf{51}\pm\textbf{3}$} & \underline{$0+1$} & {$\textbf{84.15}\pm\textbf{4.58}$}\\ \hline
\end{tabular}%
}

\label{Table-2}
\end{table*}

\begin{table*}[t]
\centering
\large
\caption{The controlled experiments to validate the effectiveness of the proposed RP-CATE architecture. The best results are in bold}
\resizebox{\textwidth}{!}{%
\begin{tabular}{c|c|c|c|c|c|
>{\columncolor[HTML]{EFEFEF}}c }
\hline

  \multicolumn{1}{c|}{\centering Model} &
  \multicolumn{1}{p{4cm}|}{\centering MAE} &
  \multicolumn{1}{p{4cm}|}{\centering RMSE} &
  \multicolumn{1}{p{4cm}|}{\centering AER(\%)} &
  \multicolumn{1}{p{4cm}|}{\centering \#Err\textless{}1\%} &
  \multicolumn{1}{p{4cm}|}{\centering \#Err\textgreater{}5\%} &
  \multicolumn{1}{p{4cm}}{\cellcolor[HTML]{EFEFEF} \centering MIR(\%)}

\\ \hline
(RP-)CATE & $0.0121 \pm 0.0314$ & $0.0220\pm0.0728$ & $1.16\pm1.41$ & $39\pm16$               & $0+9$                    & $67.81\pm19.63$ \\
RP-(CA)TE & $0.0093\pm0.0037$ & $0.0132\pm0.0033$ & $1.04\pm0.44$ & $32\pm13$               & {$\textbf{0}+\textbf{1}$}           & $61.50\pm17.48$ \\
\textbf{RP-CATE(Ours)} & {$\textbf{0.0084}\pm\textbf{0.0017}$} & {$\textbf{0.0132}\pm\textbf{0.0020}$} & {$\textbf{0.81}\pm\textbf{0.09}$} & {$\textbf{51}\pm\textbf{3}$} & {$\textbf{0}+\textbf{1}$} & {$\textbf{84.15}\pm\textbf{4.58}$} \\ \hline
\end{tabular}%
}

\label{Table-3}
\end{table*}

\subsubsection{Hyperparameters and Implementation}


RP-CATE involves four main hyperparameters: the feature \(x^{\prime}\) used to convert the input \(X\) into \(PSD\); the window size \(w\) used to transform the input data into \(PID\); the number of repetitions $N$ of the loop body in RP-CATE; and the learning rate \(lr\). Based on engineering experience, we select the most critical feature from the \(n\) available features as \(x^{\prime}\), which has the greatest impact on the prediction results. The optimal values for \(w\), $N$, and \(lr\) were found using grid search within the following ranges: \{9, 25\}, \{1, 2, 3, 4, 5\}, and \{0.01, 0.001\}, respectively. The final selections were \(w=25\), \({N}=2\), and \(lr=0.001\) and we used the Adam optimizer for training. For other baselines, we utilized the default settings from their public implementations.

\subsection{RP-CATE vs Transformer (RQ1)}
Since RP-CATE was developed based on Transformer, it is necessary to compare it with the Transformer. We randomly selected a seed to initialize the model parameters and conducted experiments with any combination of \(w=\{9, 25\}\) and \({N}=\{1, 2, 3, 4, 5\}\). The results are presented in Table \ref{Table-1}. From the table, we can see that RP-CATE outperforms the Transformer across all six metrics, regardless of window size or the number of repetitions. We speculate that the Transformer's unsatisfactory performance may be attributed to its self-attention mechanism, which might not be well-suited for industrial datasets. Consequently, the Transformer may not be entirely applicable to this modeling task. As a result, RP-CATE replaces self-attention with channel attention, and experimental results demonstrate the effectiveness of this modification. In addition, Table \ref{Table-1} illustrates the performance variations of RP-CATE with different window sizes and repetitions. We found that the model performs better overall with \(w=25\) compared to \(w=9\) and Transformer achieved the highest MIR with $w=25$, $N=1$. Furthermore, when $w=25$, the MIR values for $N=\{1, 2, 4, 5\}$ showed minimal differences. After carefully balancing model performance and complexity, we selected the hyperparameter combination of $w=25$, $N=2$. Thus, in subsequent experiments, we use \(w = 25\), $N$ = 2 as the default configuration for RP-CATE, and \(w = 25\), $N$ = 1 for the Transformer.

\subsection{Overall Performance (RQ2)}

In this part, each experiment was repeated 5 times with random initialization, with the median of these results presented in the table, and the variation of the other experimental results relative to this median was recorded. As shown in Table \ref{Table-2}, RP-CATE achieves the best results on 5/6 metrics, and decreases MAE from 0.0781 to 0.0084. In particular, according to MIR, RP-CATE demonstrates approximately an 85\% improvement compared to the mechanistic model. Moreover, after applying RP-CATE, nearly 83\% (51/60) of samples have a relative error of less than 1\%, and almost 100\% of samples have a relative error of less than 5\%. Upon closer inspection, we observe that RP-CATE exhibits the least variation in results across the five experiments compared to the median presented in the table, which indicates RP-CATE is more robust than other benchmarks. The above analysis tells the proposed RP-CATE are more efficient in modeling tasks than the other baselines.


\subsection{RP-CATE Architecture (RQ3)}
RP Module was designed to uncover underlying associations, e.g., monotonicity, between the inputs at different time steps, while the Channel Attention Module assessed the impact of different features within the input at each time step on the prediction results. To validate the effectiveness of these two core modules, we conducted the following controlled experiments: (1) (RP-)CATE, which omits the RP Module; (2) RP-(CA)TE, which excludes the Channel Attention Module; (3) RP-CATE, with the full structure. As shown in Table \ref{Table-3}, without RP Module, (RP-)CATE fails to learn the underlying associations within industrial data, resulting in a significant drop in MIR. This observation supports the initial claim that uncovering such underlying associations can enhance model predictive performance. On the other hand, the absence of Channel Attention Module in RP-(CA)TE prevents the model from considering the varying impacts of different features on prediction results, leading to a loss of flexibility and a decrease in MIR. Overall, (RP-)CATE outperforms RP-(CA)TE on the \#Err$<$1\% metric ($39\pm16$ vs $32\pm13$) but underperforms on the \#Err$>$5\% metric ($0+9$ vs $0+1$). This indicates that Channel Attention Module helps improve the upper bound of model prediction performance, while the RP Module contributes to enhancing the lower bound. In summary, the results presented in Table \ref{Table-3} robustly demonstrate the efficacy of our proposed RP-CATE architecture.

\begin{figure}[t]
\centering
\includegraphics[width=0.9\columnwidth]{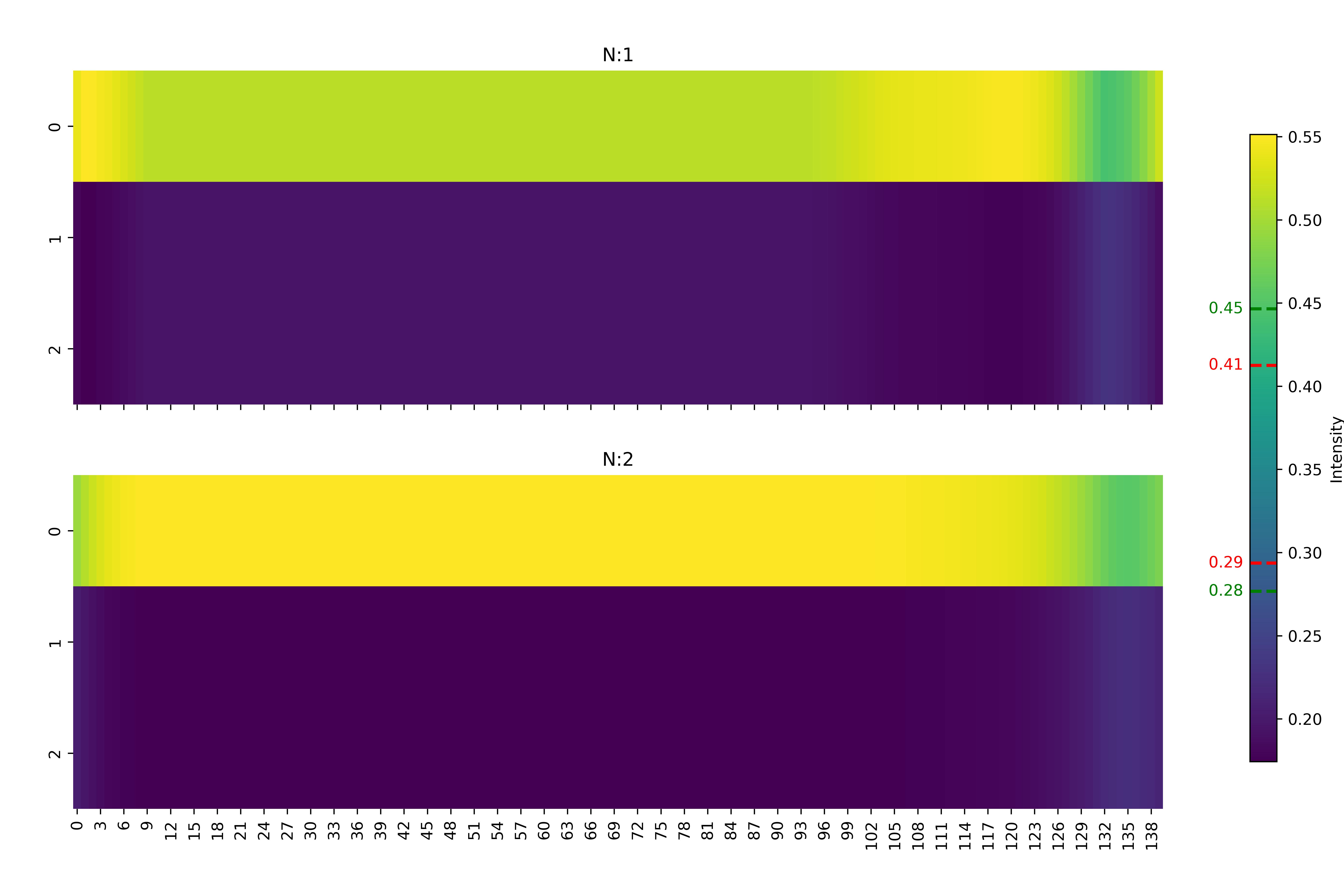} 
\caption{Attention maps. The y-axis is the feature indices of the dataset, and the x-axis is the sample indices. On the color bar to the right, the green dashed lines indicates the average attention weights for different features under N:1, while the red represents the average attention weights under N:2.}

\label{fig-4}
\end{figure}

\subsection{Visualization of Attention Maps (RQ4)}
Fig.~\ref{fig-4} illustrates the attention maps for different samples in the dataset. The attention map obtained after the first iteration of the loop body in RP-CATE is labeled as N:1, with the average attention weights for different features highlighted in red on the color bar, where the average weight for feature 0 is 0.4126, and for features 1 and 2, it is 0.2937 each. After the second iteration, the attention map is labeled as N:2, with average weights shown in green where the average weight for feature 0 is 0.4465, and for features 1 and 2, it is 0.2768 each. It is evident that feature 0 has the highest average attention weight, which has the greatest impact on the final prediction, followed by the other two features. Furthermore, we observe that the attention weights for the same feature remain nearly stable. This stability is due to the fact that in this application scenario, the three features are independent variables used to compute the target value, and the functional mapping relationship between these independent variables and the target value largely determines their attention weights. Different sample values essentially correspond to different values of the independent variables within the functional mapping relationship, which apparently does not affect the mapping itself. Therefore, the attention weights for the same feature do not vary significantly across different samples. In fact, the functional mapping relationship is essentially the mechanistic model of the modeling object.

\section{Conclusion}
This paper introduces RP-CATE, a novel architecture for machine learning-based modeling based on the Transformer framework, which can be applied to synergize with the mechanistic model to improve the industrial modeling accuracy. Firstly, we proposed PSD Module and RP Module to better capture the underlying association within the industrial datasets. Next, we introduced the Channel Attention Module, which consists of two blocks to assign attention weights to different features. Finally, the Feed-Forward Module and Prediction Module were designed to process the attention-weighted intermediate results and generate the final predictions. Experimental results demonstrate that RP-CATE not only outperforms the Transformer in modeling tasks but also exhibits cutting-edge performance compared to both conventional and advanced baseline methods. As an innovative architecture for data-driven models, RP-CATE holds potential for applications across various fields, including power systems, physics, and mathematics, suggesting promising directions for future research.

\bibliographystyle{IEEEtran}
\bibliography{Myref}

@article{lee1975generalized,
  title={A generalized thermodynamic correlation based on three-parameter corresponding states},
  author={Lee, Byung Ik and Kesler, Michael G},
  journal={AIChE Journal},
  volume={21},
  number={3},
  pages={510--527},
  year={1975},
  publisher={Wiley Online Library}
}

@article{kong2023collaborative,
  title={Collaborative extraction of intervariable coupling relationships and dynamics for prediction of silicon content in blast furnaces},
  author={Kong, Liyuan and Yang, Chunjie and Lou, Siwei and Cai, Yu and Huang, Xiaoke and Sun, Mingyang},
  journal={IEEE Transactions on Instrumentation and Measurement},
  volume={72},
  pages={1--13},
  year={2023},
  publisher={IEEE}
}

@article{zhu2022dynamic,
  title={Dynamic graph-based adaptive learning for online industrial soft sensor with mutable spatial coupling relations},
  author={Zhu, Kun and Zhao, Chunhui},
  journal={IEEE Transactions on Industrial Electronics},
  volume={70},
  number={9},
  pages={9614--9622},
  year={2022},
  publisher={IEEE}
}

@article{chen2024residual,
  title={Residual-aware deep attention graph convolutional network via unveiling data latent interactions for product quality prediction in industrial processes},
  author={Chen, Yitao and Wang, Yalin and Sui, Qingkai and Yuan, Xiaofeng and Wang, Kai and Liu, Chenliang},
  journal={Expert Systems with Applications},
  volume={245},
  pages={123078},
  year={2024},
  publisher={Elsevier}
}

@inproceedings{zhao2017time,
  title={Time-weighted LSTM model with redefined labeling for stock trend prediction},
  author={Zhao, Zhiyong and Rao, Ruonan and Tu, Shaoxiong and Shi, Jun},
  booktitle={2017 IEEE 29th international conference on tools with artificial intelligence (ICTAI)},
  pages={1210--1217},
  year={2017},
  organization={IEEE}
}

@article{banchero2018comparison,
  title={Comparison between multi-linear-and radial-basis-function-neural-network-based QSPR Models for the prediction of the critical temperature, critical pressure and acentric factor of organic compounds},
  author={Banchero, Mauro and Manna, Luigi},
  journal={Molecules},
  volume={23},
  number={6},
  pages={1379},
  year={2018},
  publisher={MDPI}
}

@article{sherstinsky2020fundamentals,
  title={Fundamentals of recurrent neural network (RNN) and long short-term memory (LSTM) network},
  author={Sherstinsky, Alex},
  journal={Physica D: Nonlinear Phenomena},
  volume={404},
  pages={132306},
  year={2020},
  publisher={Elsevier}
}

@article{mahjoub2022predicting,
  title={Predicting energy consumption using LSTM, multi-layer GRU and drop-GRU neural networks},
  author={Mahjoub, Sameh and Chrifi-Alaoui, Larbi and Marhic, Bruno and Delahoche, Laurent},
  journal={Sensors},
  volume={22},
  number={11},
  pages={4062},
  year={2022},
  publisher={MDPI}
}

@article{vaswani2017attention,
  title={Attention is all you need},
  author={Vaswani, Ashish and Shazeer, Noam and Parmar, Niki and Uszkoreit, Jakob and Jones, Llion and Gomez, Aidan N and Kaiser, {\L}ukasz and Polosukhin, Illia},
  journal={Advances in neural information processing systems},
  volume={30},
  year={2017}
}

@article{bai2018empirical,
  title={An empirical evaluation of generic convolutional and recurrent networks for sequence modeling},
  author={Bai, Shaojie and Kolter, J Zico and Koltun, Vladlen},
  journal={arXiv preprint arXiv:1803.01271},
  year={2018}
}

@article{su2019architecture,
  title={An architecture of deep learning in QSPR modeling for the prediction of critical properties using molecular signatures},
  author={Su, Yang and Wang, Zihao and Jin, Saimeng and Shen, Weifeng and Ren, Jingzheng and Eden, Mario R},
  journal={AIChE Journal},
  volume={65},
  number={9},
  pages={e16678},
  year={2019},
  publisher={Wiley Online Library}
}

@article{liu2016dislocated,
  title={Dislocated time series convolutional neural architecture: An intelligent fault diagnosis approach for electric machine},
  author={Liu, Ruonan and Meng, Guotao and Yang, Boyuan and Sun, Chuang and Chen, Xuefeng},
  journal={IEEE Transactions on Industrial Informatics},
  volume={13},
  number={3},
  pages={1310--1320},
  year={2016},
  publisher={IEEE}
}

@article{li2020novel,
  title={A novel scalable method for machine degradation assessment using deep convolutional neural network},
  author={Li, Pin and Jia, Xiaodong and Feng, Jianshe and Zhu, Feng and Miller, Marcella and Chen, Liang-Yu and Lee, Jay},
  journal={Measurement},
  volume={151},
  pages={107106},
  year={2020},
  publisher={Elsevier}
}

@article{chen2020framework,
  title={A framework of hybrid model development with identification of plant-model mismatch},
  author={Chen, Yingjie and Ierapetritou, Marianthi},
  journal={AIChE Journal},
  volume={66},
  number={10},
  pages={e16996},
  year={2020},
  publisher={Wiley Online Library}
}

@phdthesis{abakporo2021general,
  title={General Thermodynamic Framework for Organic Rankine Cycles},
  author={Abakporo, Obiechina I},
  year={2021},
  school={Florida Agricultural and Mechanical University}
}

@article{borowski2021digitization,
  title={Digitization, digital twins, blockchain, and industry 4.0 as elements of management process in enterprises in the energy sector},
  author={Borowski, Piotr F},
  journal={Energies},
  volume={14},
  number={7},
  pages={1885},
  year={2021},
  publisher={MDPI}
}

@article{sansana2021recent,
  title={Recent trends on hybrid modeling for Industry 4.0},
  author={Sansana, Joel and Joswiak, Mark N and Castillo, Ivan and Wang, Zhenyu and Rendall, Ricardo and Chiang, Leo H and Reis, Marco S},
  journal={Computers \& Chemical Engineering},
  volume={151},
  pages={107365},
  year={2021},
  publisher={Elsevier}
}

@book{velten2024mathematical,
  title={Mathematical modeling and simulation: introduction for scientists and engineers},
  author={Velten, Kai and Schmidt, Dominik M and Kahlen, Katrin},
  year={2024},
  publisher={John Wiley \& Sons}
}

@article{scheidegger2018introductory,
  title={An introductory guide for hybrid simulation modelers on the primary simulation methods in industrial engineering identified through a systematic review of the literature},
  author={Scheidegger, Anna Paula Galv{\~a}o and Pereira, T{\'a}bata Fernandes and de Oliveira, Mona Liza Moura and Banerjee, Amarnath and Montevechi, Jos{\'e} Arnaldo Barra},
  journal={Computers \& Industrial Engineering},
  volume={124},
  pages={474--492},
  year={2018},
  publisher={Elsevier}
}

@article{carter2023review,
  title={Review of interpretable machine learning for process industries},
  author={Carter, A and Imtiaz, S and Naterer, GF},
  journal={Process Safety and Environmental Protection},
  volume={170},
  pages={647--659},
  year={2023},
  publisher={Elsevier}
}

@article{elsheikh2023control,
  title={Control of an industrial distillation column using a hybrid model with adaptation of the range of validity and an ANN-based soft sensor},
  author={Elsheikh, Mohamed and Ortmanns, Yak and Hecht, Felix and Ro{\ss}mann, Volker and Kr{\"a}mer, Stefan and Engell, Sebastian},
  journal={Chemie Ingenieur Technik},
  volume={95},
  number={7},
  pages={1114--1124},
  year={2023},
  publisher={Wiley Online Library}
}

@article{mcbride2020hybrid,
  title={Hybrid semi-parametric modeling in separation processes: a review},
  author={McBride, Kevin and Sanchez Medina, Edgar Ivan and Sundmacher, Kai},
  journal={Chemie Ingenieur Technik},
  volume={92},
  number={7},
  pages={842--855},
  year={2020},
  publisher={Wiley Online Library}
}

@article{elman1990finding,
  title={Finding structure in time},
  author={Elman, Jeffrey L},
  journal={Cognitive science},
  volume={14},
  number={2},
  pages={179--211},
  year={1990},
  publisher={Wiley Online Library}
}

@article{bertolini2021machine,
  title={Machine Learning for industrial applications: A comprehensive literature review},
  author={Bertolini, Massimo and Mezzogori, Davide and Neroni, Mattia and Zammori, Francesco},
  journal={Expert Systems with Applications},
  volume={175},
  pages={114820},
  year={2021},
  publisher={Elsevier}
}

@article{sarker2022ai,
  title={AI-based modeling: techniques, applications and research issues towards automation, intelligent and smart systems},
  author={Sarker, Iqbal H},
  journal={SN Computer Science},
  volume={3},
  number={2},
  pages={158},
  year={2022},
  publisher={Springer}
}

@article{adewumi2013acentric,
  title={Acentric Factor and Corresponding States},
  author={Adewumi, Michael},
  journal={Pennsylvania State University. Retrieved},
  pages={11--06},
  year={2013}
}

@article{boublia2023critical,
  title={Critical properties of ternary deep eutectic solvents using group contribution with extended lee--kesler mixing rules},
  author={Boublia, Abir and Lemaoui, Tarek and Almustafa, Ghaiath and Darwish, Ahmad S and Benguerba, Yacine and Banat, Fawzi and AlNashef, Inas M},
  journal={ACS omega},
  volume={8},
  number={14},
  pages={13177--13191},
  year={2023},
  publisher={ACS Publications}
}

@article{sarker2021machine,
  title={Machine learning: Algorithms, real-world applications and research directions},
  author={Sarker, Iqbal H},
  journal={SN computer science},
  volume={2},
  number={3},
  pages={160},
  year={2021},
  publisher={Springer}
}

@inproceedings{xiao2017modeling,
  title={Modeling the intensity function of point process via recurrent neural networks},
  author={Xiao, Shuai and Yan, Junchi and Yang, Xiaokang and Zha, Hongyuan and Chu, Stephen},
  booktitle={Proceedings of the AAAI conference on artificial intelligence},
  volume={31},
  number={1},
  year={2017}
}



\end{document}